\pdfoutput=1 

\documentclass[journal]{IEEEtran}

\usepackage{cite}
\usepackage{amsmath,amssymb,amsfonts}
\usepackage{algorithmic}
\usepackage{graphicx}
\usepackage{textcomp}
\usepackage{epstopdf,url,color,array}
\usepackage{multirow} % for multirow format in tables
\usepackage[ruled,vlined]{algorithm2e}
\usepackage{tablefootnote}

\ifCLASSINFOpdf
\else
\fi
\begin{document}
%\title{Discriminative information learning for breast cancer pathology image classification via deep convolutional autoencoder}
\title{Discriminative Pattern Mining for Breast Cancer Histopathology Image Classification via Fully Convolutional Autoencoder}

\author{Xingyu~Li,~\IEEEmembership{Member,~IEEE,} Marko Radulovic, Ksenija Kanjer,
        and~Konstantinos~N.~Plataniotis,~\IEEEmembership{Fellow,~IEEE}% <-this % stops a space
\thanks{Xingyu Li (xingyu.li@mail.utoronto.ca) and Konstantinos N. Plataniotis (kostas@ece.utoronto.ca) are with Multimedia Lab, The Edward S. Rogers Department of Electrical and Computer Engineering, University of Toronto, Toronto, Ontario, Canada.

Marko Radulovic and Ksenija Kanjer are with the National Cancer Research Centre, Department of Experimental Oncology, Institute for Oncology and Radiology, Belgrade, Serbia.

\copyright 20xx IEEE. Personal use of this material is permitted. Permission
from IEEE must be obtained for all other uses, in any current or future
media, including reprinting/republishing this material for advertising or
promotional purposes, creating new collective works, for resale or
redistribution to servers or lists, or reuse of any copyrighted
component of this work in other works.}% <-this % stops a space
}

\maketitle

\begin{abstract}
Accurate diagnosis of breast cancer in histopathology images is challenging due to the heterogeneity of cancer cell growth as well as of a variety of benign breast tissue proliferative lesions. In this paper, we propose a practical and self-interpretable invasive cancer diagnosis solution. With minimum annotation information, the proposed method mines contrast patterns between normal and malignant images in unsupervised manner and generates a probability map of abnormalities to verify its reasoning. Particularly, a fully convolutional autoencoder is used to learn the dominant structural patterns among normal image patches. Patches that do not share the characteristics of this normal population are detected and analyzed by one-class support vector machine and 1-layer neural network. We apply the proposed method to a public breast cancer image set. Our results, in consultation with a senior pathologist, demonstrate that the proposed method outperforms existing methods. The obtained probability map could benefit the pathology practice by providing visualized verification data and potentially leads to a better understanding of data-driven diagnosis solutions.

%Consider the case of a medical alert system that should notify staff when combinations of measurements (lab results, sensor data, etc.) are unusual and thus suspect.All patients that do not share the characteristics of this normal population are at higher risk of developing metastases
\end{abstract}


\IEEEpeerreviewmaketitle

\section{Introduction}\label{sec:introduction}
Breast cancer is the second most common cancer in women. Invasive, malignant properties of breast cancer cell growth contribute to poor patient prognosis \cite{Dillon2010}, and dictate precise early diagnosis and treatment, with an aim to reduce breast cancer morbidity rate. In this study, we particularly focus on the qualification of risky, aggressive characteristics of breast histomorphological patterns, as one of the basic features of invasiveness of breast carcinoma.

%\begin{figure*}
%	\centering
%		\includegraphics[width=7in]{image/fig1}
%		\centering	\caption{Examples of hemotoxylin and eosin stained images for (a) normal breast tissue and (b) invasive carcinoma with a magnification of $40\times$. The left image corresponds to a normal tissue where normal epithelial cells lie on the membrane of ductulo-lobular structures; while in the right image malignant cells invade and spread into surrounding tissue.}
%	\label{fig:imgdemo}
%\end{figure*}

\begin{figure}
	\centering
		\includegraphics[width=85mm]{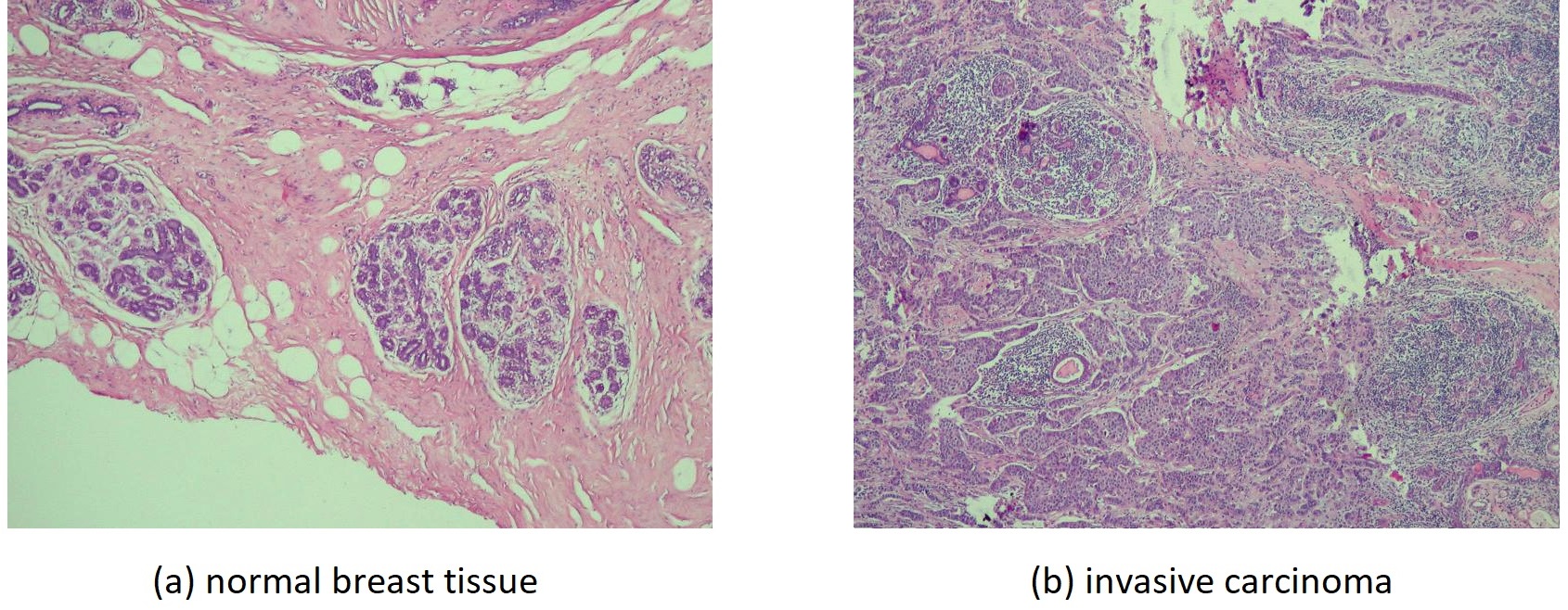}
		\centering	\caption{Examples of hemotoxylin and eosin stained images for (a) normal breast tissue and (b) invasive carcinoma with a magnification of $40\times$. The left image corresponds to a normal tissue where normal epithelial cells lie on the membrane of ductulo-lobular structures; while in the right image malignant cells invade and spread into surrounding tissue.}
	\label{fig:imgdemo}
\end{figure}

With the advance of imaging device and machine learning technology, digital {\color{black}histopathology} image analysis becomes a promising approach to consistent and cost-efficient cancer diagnosis. Particularly for invasive breast cancer, based on the common knowledge that cancerous cells break through the basement membrane of ductulo-lobular structures and infiltrate into surrounding tissues - the feature of invasiveness  \cite{Robertson2017} (as shown in Fig. \ref{fig:imgdemo}), many algorithms were proposed to classify breast {\color{black}histopathology} images using nuclei's morphology and spatial-distribution features \cite{Veta2014}. In literature, the most common solution to breast cancer image diagnosis is to train a classifier in a supervised learning manner. Then handcrafted features of a query image are passed to the trained algorithm for a yes/no label \cite{Filipczuk2013,George2014,Kandemir2014, Bhandari2015,Li2015, XLi2016,Spanhol2016a}. With the success of deep learning, data-driven methods, especially the end-to-end training of convolutional neural network, are adopted more often in recent breast cancer {\color{black}histopathology} image classification studies \cite{Spanhol2016b,Araújo2017,Spanhol2017}. Though breast cancer image diagnosis has achieved impressive progress, the issue of self-interpretability in existing diagnosis approaches is less addressed. {\color{black}Self-interpretability refers to the capability of an approach to explain and verify its reasoning and results.} Without self-interpretability, attempts to improve {\color{black}histopathology} image diagnosis is prone to be limited to trial-and-error.

%\Figure[t!](topskip=0pt, botskip=0pt, midskip=0pt)[width=85mm]{image/fig1}
%{Examples of hemotoxylin and eosin stained images for (a) normal breast tissue and (b) invasive carcinoma with a magnification of $40\times$. The left image corresponds to a normal tissue where normal epithelial cells lie on the membrane of ductulo-lobular structures; while in the right image malignant cells invade and spread into surrounding tissue.\label{fig:imgdemo}}

To address the self-interpretability issue in breast cancer pathology image diagnosis, one solution is to generate labels for image pixels or small image patches in order to infer locations of suspected abnormalities in a query image. To this end, several studies made efforts to classify small image patches or pixels via supervised/semi-supervised learning  \cite{Cruz-Roa2014,Sirinukunwattana2016,Cruz-Roa2017,Bidar2018,Peikari2018}. It should be noted that these solutions require a large amount of images manually-annotated at the image pixel level. {\color{black}Due to the complexity and time-intensive properties of pathological annotations and privacy concerns in clinical practice, sufficient amount of well-labeled patches are difficult to collect.}% and access. In addition, normal tissue samples are greatly more than abnormal cases in pathology, and the resulting unbalanced classes may also degrade performance of supervised learning.}% Due to difficulty and time-consuming nature of pathology annotation, databases with pixel annotation are expensive to collect and access.}
%On the other hand, training a detector to identify malignant tummor in {\color{black}histopathology} images usually requires a large amount of manually-annotated images by pathologists. Since the annotation of {\color{black}histopathology} images is time-consuming and costly, it is not easier to collect very large image sets. 

This study attempts to tackle the self-interpretability issue in breast cancer diagnosis and presents a novel convolutional autoencoder-based contrast pattern mining approach to detect the invasive component of malignant breast epithelial growth in routine hematoxylin and eosin (H\&E) stained {\color{black}histopathology} images. As opposed to prior studies that require image sets with pixel annotation, our method requires only image labels as the minimal prior knowledge in training. By mining dominant patterns in images of normal breast tissues, the method generates a probability map to infer locations of abnormalities in an image. As a pathology image may contain both normal and cancerous tissues, the proposed method divides an image into small patches to facilitate local characteristics learning. It should be noted that due to the lack of pixel annotation indicating the locations of abnormal cell growth patterns in images, this problem is very challenging in two folds. 
\begin{enumerate}
	\item The algorithm is expected to learn contrast patterns between normal and malignant/invasive growth based on the knowledge of image labels. Effective differentiation between normal and abnormal histomorphology via unsupervised learning is the key issue for the correct identification of cancerous growth.
	\item As a {\color{black}histopathology} image may contain both normal and cancerous tissue, labels of local patches may be inconsistent with the known image label. The method needs to learn a mapping function between local patches and image labels.
\end{enumerate}

Note that though we {\color{black}do not} know whether patches from images labeled as malignant really contain malignant cells/structures, patches from normal images do not contain cancerous cells certainly. So, we name a patch from a normal image "true-normal" in this paper. Our original approach learns patterns in true-normal patches first and then assigns a normal/malignant label to a patch which resembles/deviates from those true normal ones. Intuition behind this originality is that in pathology, malignant cells and their growth patterns are diagnosed and graded by how different these cells are to normal cells. Specifically, to address the first challenge, we exploit the data-specific property of autoencoder (AE) networks  \cite{Bengio2013,Ponti2017} and innovate to train an under-complete deep fully convolutional AE using small patches from pathology images annotated as normal. Since the network learns local patterns in true-normal patches only, its performance degrades when the input instance is different from training patches. Hence, autoencoder's reconstruction residue suggests the similarity between the query instance and normal cases. It is noteworthy that different from standard autoencoders targeting to minimize mean square error (MSE) between input and output training instances, the proposed method trains the deep net by optimizing the structural similarity (SSIM) index \cite{Wang2004}, which enforces the network to learn the contrast and structural patterns in true-normal patches. In this study, the trained AE network is treated as a pattern mining and representation method and then combined with downstream classifiers to identify whether an image patch contains malignant cells.

To tackle the second challenge which is to infer whether a local patch contains morphological abnormalities derived by malignant cell growth, we cast the problem into the anomaly detection scenario, and introduce a novel malignant patch detector to distinguish patches containing cancerous cells from the normal ones. Particularly, the proposed detector makes the use of one class support vector machine (SVM) \cite{Schölkopf2001} to identify regions occupied by true-normal patches in the feature space and assigns abnormal labels to patches whose numerical features are located outside of the detected normal regions. Taking into account the obtained patch labels, the problem of breast cancer image classification with localization of abnormality areas is simplified to a patch-based supervised learning problem. Finally, a 1 layer neural network (NN) is trained to infer the existence of malignant tumor in a patch, followed by the generation of a probability map of abnormality in the query image. 

%In summary, the novelty of this approach is in the capability to simultaneously perform contrast pattern mining for image classification and generate visualized verification data. Specifically, the proposed diagnosis approach learns discriminative patterns in unsupervised manner from  histopathology images and interprets its results via inferring locations of abnormalities in an image. The obtained probability map could benefit the pathology practice by providing visualized data, which contributes to a better understanding of data-driven diagnostic solutions. To the best of our knowledge, our work constitutes the first attempt in literature to tackle the self-interpretability issue in histopathology image classification. 
{\color{black}In summary, this study proposes a practical, generalizable, and self-interpretable solution to pathology image based cancer diagnosis. With the minimal prior knowledge on whose-slide-imaging (WSI) labels which can be easily acquired in clinic practice, the proposed method learns discriminative patterns in unsupervised manner from histopathology images and explains its diagnosis results via inferring locations of abnormalities in an image. It is noteworthy that the proposed method is very user-friendly to pathologists, as the obtained abnormality map helps pathologists to understand and verify how machines make decisions. To the best of our knowledge, our work constitutes the first attempt in literature to tackle the self-interpretability issue in histopathology image classification.}

The rest of this paper is organized as follows. Section \ref{sec:background} provides brief introduction of machine learning techniques exploited in the proposed method and the public breast cancer biopsy image set used in this study. The problem's formal statement with notations and implementation details are presented in Section \ref{sec:formulation} and Section \ref{sec:implementation}, respectively. Experimental results and discussions are presented in Section \ref{sec:simulation}, followed by conclusions in Section \ref{sec:conclusion}.

\section{Background} \label{sec:background}
{\color{black}In this section, we will first introduce notations used in this study in Table \ref{tab:notation}. Then brief description of fully convolutional autoencoder and one-class support vector machine is presented, followed by information on the public image set used to evaluate the proposed method in this study.}

\begin{table}	
{\color{black}
\caption{Table of Notations. } 
	\centering
	\begin{tabular}{ l  l  }
	\hline 
	 Notations & Explanations \\ \hline	\hline
		$\mathcal A$, $\mathcal B$ & Trainable parameters of 1-layer NN \\ \hline
		%$\mathcal B$ & Trainable parameters of 1-layer NN \\ \hline
		$\mathcal D()$ & Decorder of AE \\ \hline
		$\mathcal E()$ & Encoder of AE \\ \hline
		$\mathcal F()$ & Patch labeling function \\ \hline
		$\mathcal G()$ & Decision function of one-class SVM \\ \hline
		$\mathcal I$ & Histopathology image \\ \hline
		$\mathcal L$ & Loss function of AE \\ \hline
		$M$ & Number of normal images \\ \hline
		$N$ & Number of malignant images \\ \hline
		$T$ & Number of training patches \\ \hline		
		$\textit c$, $\nu$ & Hyper-parameter of one-class SVM \\ \hline
		$\textit k()$ & Gaussian kernel \\ \hline
		$\textit w_{i}$, $\textit v_{i}$ & Trainable parameteres of AE \\ \hline
		$\textit x$ & Input of AE (i.e. greay-scale true-normal patches) \\ \hline
	 $\tilde{\textit x}$ & Output of AE \\ \hline
	$\tilde{\textit x}$ & residue of AE's reconstruction \\ \hline
	$\textit y$ & Patch labels \\ \hline
	 $z$ & Input of one-class SVM \\ \hline
	$\mathcal y$ & Histopathology image Label \\ \hline
	$\alpha, \beta, \gamma$ & Hyper-parameters of SSIM \\ \hline
	$\delta()$ & Dirac delta function\\ \hline
	$\lambda$ & Lagrangian multiplier of one-class SVM \\ \hline
	\end{tabular}	
	\label{tab:notation}}
\end{table}

\subsection{Fully Convolutional AE Networks}
Fully convolutional network is defined as the neural network composed of convolutional layers without any fully-connected layers \cite{Long2015}. It learns representations and makes decisions based on local spatial knowledge only. Because of its efficient learning, fully convolutional net has been popular in many image-to-image inference tasks, e.g. semantic segmentation.

Fully convolutional autoencoder is one instance of fully convolutional neural networks. The net takes input of arbitrary size and produces corresponding-sized output. Specifically, it encodes an image data $\textit x$ of arbitrary size into a low-dimensional representation $\hat{\textit x}$ such that the important properties of the original data can be reconstructed and maintained in the output $\tilde{\textit x}$. Mathematically, a fully convolutional autoencoder is composed of an encoder $\mathcal E()$ and a decorder $\mathcal D()$, each of which is a composition of a sequence of $C$ layers, i.e.
\begin{eqnarray}
	 \tilde{\textit x} &=& \mathcal D(\hat{\textit x};\textit v_1,...,\textit v_C) \\
										&=& \mathcal D(\mathcal E(\textit x;\textit w_1,...,\textit w_C);\textit v_1,...,\textit v_C), \nonumber
\end{eqnarray}
where $\tilde{\textit x}=\mathcal D(\hat{\textit x};\textit v_1,...,\textit v_C) = \mathcal D_{C}(\cdot;\textit v_{C})\circ \cdots \circ \mathcal D_{1}(\hat{\textit x};\textit v_{1})$ and $\hat{\textit x} = \mathcal E(\textit x;\textit w_1,...,\textit w_C) = \mathcal E_{C}(\cdot;\textit w_{C}) \circ \cdots \circ \mathcal E_{1}(\textit x;\textit w_1)$. $\mathcal D() \circ \mathcal E(\textit x) = \mathcal D(\mathcal E(\textit x))$ and $\textit w_{i}$ and $\textit v_{i}$ are the weights and bias for the $i^{th}$ encoder layer $\mathcal E_{i}()$ and decoder layer $\mathcal D_i()$, respectively. Conventionally, $\mathcal E_{i}()$ performs one of the following operations: a) convolution with a bank of filters, b) downsample by spatial pooling, and c) non-linear activation; and $\mathcal D_i()$ takes actions including: d) convolution with a bank of deconvolution filters, e) upsample by interpolations, and f) non-linear activation. Given a set of $T$ training sample $\{\textit x_1,...,\textit x_{T}\}$, the parameter set of autoencoder $\{\textit w_{k},\textit v_{k}, 0<k\leq C\}$ is optimized such that reconstruction $\tilde{\textit x}$ resembles input $\textit x$:
\begin{eqnarray}
 \arg\min_{\textit w_k,\textit v_k, 0<k\leq C} \frac{1}{T}\sum_{i=1}^{T}\mathcal L(\textit x_i,\tilde{\textit x_i}),
\end{eqnarray}
where $\mathcal L$ is a loss function measuring the similarity between $\textit x_i$ and $\tilde{\textit x_i}$, e.g. MSE.

\subsection{One-class Support Vector Machine}
One-class SVM is an approach for semi-supervised anomaly detection. It models the normal data as a single class that occupies a dense subset of the feature space corresponding to the kernel and aims to find the "normal" regions. A test instance that resides in such a region is accepted by the model whereas anomalies are not \cite{Schölkopf2001}. That is, it returns a function for input $z$ that takes the value +1 in the small region capturing most of normal points, and -1 elsewhere. With the training set $\{\textit z_1,...,z_{T}\}$, the duel problem of the one-class SVM solution can be formulated by
\begin{eqnarray}
	&&\min_{\alpha_i,0<i\leq T} \frac{1}{2}\sum_{i,j=0}^{T}\lambda_i \lambda_j \textit k(\textit z_i,\textit z_j) \\
	&\text{s.t.}& 0<\lambda_i \leq \frac{1}{\nu T}, \sum_{i=0}^{T}\lambda_i=1, 
\end{eqnarray}
where $\lambda_i$ is a Lagrangian multiplier for sample $\textit z_{i}$, $\textit k(\textit z_i,\textit z_j)=e^{-\| \textit z_i-\textit z_j\|^{2}/\textit c}$ is the Gaussian kernel with parameter $\textit c$, and $\nu \in (0,1]$ is a hyper-parameter that controls training errors in this optimization problem \cite{Schölkopf2001}. The samples $\{\textit z_i:\lambda_i > 0\}$ are support vectors which lay on the "normal" region boundary. For a new point $\textit z$, SVM computes the corresponding decision function $\mathcal G(\textit z) = \sum_{i=0}^{T} \lambda_i \textit k(\textit z_i,\textit z)-\rho$ and the label of the new point, $y$, is evaluated by the function's sign, i.e.
\begin{eqnarray}
	&y=\text{sgn}(\mathcal G(\textit z) = \text{sgn}\large(\sum_{j=0}^{T} \lambda_j \textit k(\textit z_j,\textit z)-\rho \large),& \\
	&\text{where } \rho = \sum_{j=0}^{T} \lambda_j \textit k(\textit z_j,\textit z_i), \forall \textit z_i:\lambda_i > 0.&
	\label{eqn:rho}
\end{eqnarray}

It should be noted that performance of one-class SVM strongly depends on the settings of their hyper-parameters $\nu$ and $\textit c$ \cite{Ghafoori2018}. However, these two parameters are application-dependent, and their settings in an efficient and unsupervised manner is still an open research problem \cite{Liu2014}.

\subsection{Breast Cancer Biopsy Image Set}

The breast cancer benchmark biopsy dataset collected from clinical samples was published by the Israel Institute of Technology \cite{imageset}. The image set consists of 361 samples, of which 119 were classified by a pathologist as normal tissue, 102 as carcinoma in situ, and 140 as invasive carcinoma. The samples were generated from patients' breast tissue biopsy slides, stained with H\&E. They were photographed using a Nikon Coolpix 995 attached to a Nikon Eclipse E600 at magnification of $40\times$ to produce images with resolution of about 5$\mu m$ per pixel. No calibration was made, and the camera was set to automatic exposure. The images were cropped to a region of interest of $760\times 570$ pixels and compressed using the lossy JPEG compression. The resulting images were again inspected by a pathologist in the Institute to ensure that their quality was sufficient for diagnosis.

\begin{figure*}
	\centering
		\includegraphics[width=7in]{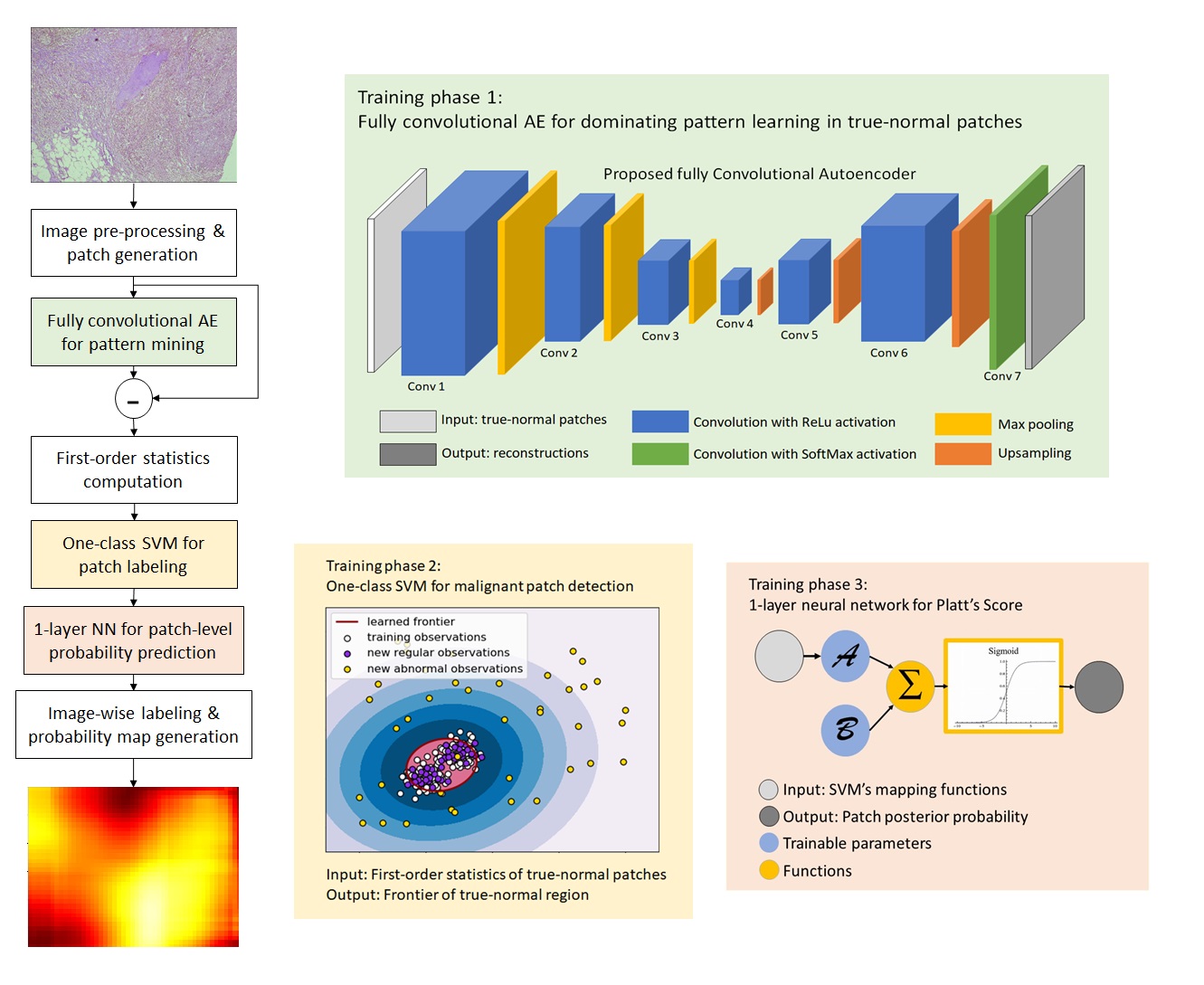}
		\centering	\caption{Overview of the proposed contrast pattern mining method for invasive breast cancer diagnosis in {\color{black}histopathology} images. The output is a probability map of malignant cell clusters in the query image, bright pixel values representing high probability. The whole system is mainly composed of three learning phases. First, a fully convolutional AE is applied to true-normal patches, so that the common patterns shared by true normal patches are learned. Based on AE's reconstruction residues, we propose to use one-class SVM to learn the regions taken by true-normal patches in the feature space. Finally, the distance to the normal region boundary in the feature space is feed to a 1-layer NN for posterior probability prediction. }
	\label{fig:Overview}
\end{figure*}

\section{Methodology} \label{sec:formulation}

%\Figure[t!](topskip=0pt, botskip=0pt, midskip=0pt)[width=170mm]{image/overview}
%{Overview of the proposed contrast pattern mining method for invasive breast cancer diagnosis in {\color{black}histopathology} images. The output is a probability map of malignant cell clusters in the query image, bright pixel values representing high probability. The whole system is mainly composed of three learning phases. First, a fully convolutional AE is applied to true-normal patches, so that the common patterns shared by true normal patches are learned. Based on AE's reconstruction residues, we propose to use one-class SVM to learn the regions taken by true-normal patches in the feature space. Finally, the distance to the normal region boundary in the feature space is feed to a 1-layer NN for posterior probability prediction. \label{fig:Overview}}

\subsection{System Overview}
Given a dataset $\{(\mathcal I_{1}, \mathcal y_{1}),(\mathcal I_{2},\mathcal y_{2}),...,(\mathcal I_{K},\mathcal y_{K})\}$ with $K$ samples, where $\mathcal I_{i}$ is an image and $\mathcal y_{i}\in \{-1,1\}$ is a class label indicating whether the corresponding image contains malignant tumor, the goal is to predict the label $\mathcal y$ for an query image $\mathcal I$ and at the same time to generate a probability map indicating suspected abnormal regions in image $\mathcal I$. For simplification, $\mathcal I_{i}^{-}$ and $\mathcal I_{i}^{+}$ denote that $\mathcal I_{i}$ is a normal or malignant image in following sections, respectively. Without loss of generality, assume that there are $N$ normal images and $M$ invasive breast cancer images in the dataset, $0<N,M<K$ and $N+M=K$, and the normal images are ordered before the malignant ones. That is, the training set is organized as $\{\mathcal I_{1}^{-},\mathcal I_{2}^{-},...,\mathcal I_{N}^{-},\mathcal I_{N+1}^{+},\mathcal I_{N+M-1}^{+},\mathcal I_{K}^{+}\}$. 

To generate the probability map of cancerous cells, we propose a patch-based learning solution, whose schematic diagram is depicted in Fig. \ref{fig:Overview}. Specifically, for each training image $\mathcal I_{i}$, we extract $l_{i}$ overlapping image patches, denoted by $\{\textit x_{i1},...,\textit x_{i,l_{i}}\}$. Patches from normal image $\mathcal I_{i}^{-}$ are assigned label $\textit y_{i,j}=-1$. However, since patches from malignant image $\mathcal I_{i}^{+}$ may contain normal tissues only, patch labels $\textit y_{i,j}$ are unknown with a positive constraint that at least one patch contains cancerous cells, i.e. $\max y_{i,j}= 1$ for $0<j\leq l_i$. If we collect all image patches into a patch set $\{(\textit x_{11},\textit y_{11}),...,(\textit x_{i,j},\textit y_{i,j}),...,(\textit x_{K,l_{K}},\textit y_{K,l_{K}})\}$, then it is evident that in the total $T=T_N+T_M$ patches, the first $T_N=\sum_{i=1}^{N}l_{i}$ patches are true-normal ones from the normal {\color{black}histopathology} images while the remaining $T_M=\sum_{i=N+1}^{K}l_{i}$ patches are from malignant images.

In the training phase, the target is to learn a mapping function $\mathcal F: \textit x \rightarrow \{-1,+1\}$ from the training set $\{(\textit x_{i,j},\textit y_{i,j}),0<j\leq l_i, 0<i\leq K\}$. Since labels of the last $T_M$ patches from malignant images $\mathcal I_{i}^{+}$ are unavailable, we make the use of unsupervised learning methods for discriminative patterns mining to classify image patches. %executed mainly based on the true-normal patches $\textit x_{i}$ for $0<i\leq T_n$. 
As shown in Fig. \ref{fig:Overview}, an under-complete deep convolutional autoencoder and a one-class SVM, both trained with true-normal patches, are used to implicitly mine dominant patterns in true-normal patches. As a representation method, autoencoder learns the common information in training patches and delivers contrast patterns in its reconstruction residues. Briefly, a normal patch has a low construction error while a malignant patch is with a high residue. Then the trained one-class SVM assigns a label $\{+1,-1\}$ to patch $\textit x_{i,j}$ from malignant image $\mathcal I^{+}_{i}$ based on autoencoder's residues. Since the one-class SVM is incapable of generating a probability value, with the obtained decision function and labels generated by the SVM, a 1-layer NN is trained to obtain Platt's score \cite{Platt1999} as patch-based posterior probabilities.

In the testing phase, $l$ overlapping patches $\textit x_{j}$ for $0<j\leq l$ are extracted from the query image $\mathcal I$. The learnt mapping function $\mathcal F$, achieved by the trained autoencoder and the one-class SVM with the 1-layer NN, is operated on each patch, generating a patch label and a value between $[0,1]$ indicating the probability that the patch contains malignant tumor. Finally, image classification and a probability map are inferred from obtained patch labels.

\subsection{Contrast Pattern Mining via Convolutional Autoencoder}

Though cell's spatial distribution is the one of the key features for invasive breast cancer diagnosis, this feature is not trivial to quantify. This is because specific structures and patterns of malignant cell clusters differ very much among different tumors and also locally within the same tumor. The incompleteness of local patch labels in this study makes the problem more challenging. Hence in this study, a data-driven solution, specifically, deep convolutional autoencoder, is used to learn the contrast patterns in the training data. 

{\color{black}It is noteworthy that an autoencoder is used as a generator of normal patches in this study. In pathology, normal breast tissues usually share certain common patterns, whereas abnormal patterns are highly heterogeneous and features learned from limited quantity of malignant samples may not be descriptive for unseen samples. To overcome this challenge, our method proposes detection of histological abnormalities implicitly by identifying the common patterns in normal breast images. To this end, we make use of the data-specific property of autoencoder and train an autoencoder to learn histological knowledge in true-normal patches.}

%\subsubsection{Deep Convolutional Autoencoder}
\subsubsection{Architecture}
Since {\color{black}histopathology} images are H\&E stained and image patches from normal and malignant biopsy images share certain common features, efforts are made to enforce the autoencoder to learn discriminative structural patterns via designing autoencoder's architecture. Particularly in this study, the experimental images from the Israel Institute of Technology image set \cite{imageset} have a magnification of 40x where pixel size is approximately 5$\mu m$. Since the diameters of breast epithelial cells' nuclei stained by H\&E are approximately  6$\mu m$ \cite{Araújo2017}, nuclei radii are between 1 and 3 pixels. Thus, we design the proposed convolutional autoencoder whose encoder $\mathcal E()$ and decorder $\mathcal D()$ both are with $C=6$ and have 3 convolutional layers, such that the nuclei-scale features, nuclei organization features, and the tissue structural-scale features are explored. Table \ref{tab:Autoencoder} provides detailed architecture of the proposed autoencoder and associates histological features with network layers. Note that the first 6 convolutional layers are composed of the Rectified Linear (Relu) activation unit $Relu(x) = \max(0,x)$ \cite{AlexNet2012}. We select the sigmoid function $sigmoid(x)=\frac{1}{1+e^{-x}}$ as the activation function in the last convolutional layer to generate a grayscale image in the range of $[0,1]$.

\begin{table*}	
\caption{Architecture of the proposed convolutional autoencoder} 
	\centering
	\begin{tabular}{ |c | c | c | c | c | c | c|}
	\hline 
	\multirow{2}{*}{}& \multirow{2}{*}{Layer} & \multirow{2}{*}{Layer type} & \multirow{2}{*}{Filter Size} & \multirow{2}{*}{Activation} & \multirow{2}{*}{Input/Output Dimension} & Histological association with breast  \\ %\cline{1-6}
	& &  &  & &  &  cancer images in magnification of 40x \\ \hline
	\multirow{7}{*}{Encoder $\mathcal E()$ }& 0 & input & &  & $256\times 256 \times 1$ &    \\ \cline{2-7} 
	& 1 & Convolutional & $3\times 3 \times 16 $ & Relu & $256\times 256 \times 16$ &  \multirow{2}{*}{nuclei \& edge}  \\ \cline{2-6} 
	& 2 & Max-polling & $2\times 2 $ &  & $128\times 128 \times 16$ &    \\ \cline{2-7}
	& 3 & Convolutional & $3\times 3 \times 8$ & Relu & $128\times 128 \times 8$ &  \multirow{2}{*}{nuclei organization}  \\ \cline{2-6} 
	& 4 & Max-polling & $2\times 2 $ &  & $64\times 64 \times 8$ &    \\ \cline{2-7} 
	& 5 & Convolutional & $3\times 3 \times 8$ & Relu & $64\times 64 \times 8$ &  \multirow{4}{*}{Structure \& tissue organization}  \\ \cline{2-6} 
	& 6 & Max-polling & $2\times 2 $ &  & $32\times 32 \times 8$ &    \\ \cline{1-6}
	\multirow{7}{*}{Decoder $\mathcal D()$ } & 7 & Convolutional & $3\times 3 \times 8$ & Relu & $32\times 32 \times 8$ &    \\ \cline{2-6} 
	& 8 & Upsampling & $2\times 2 $ &  & $64\times 64 \times 8$ &    \\ \cline{2-7}
	& 9 & Convolutional & $3\times 3 \times 8$ & Relu & $64\times 64 \times 8$ &  \multirow{2}{*}{nuclei organization}  \\ \cline{2-6} 
	& 10 & Upsampling & $2\times 2 $ &  & $128\times 128 \times 8$ &    \\ \cline{2-7}
	& 11 & Convolutional & $3\times 3 \times 16$ & Relu & $128\times 128 \times 16$ &  \multirow{2}{*}{nuclei \& edge}  \\ \cline{2-6} 
	& 12 & Upsampling & $2\times 2 $ &  & $256\times 256 \times 16$ &    \\ \cline{2-7}   
	& 13 & Convolutional/Output & $3\times 3 \times 1$ & Sigmoid & $256\times 256 \times 1$ &    \\ \hline 
	\end{tabular}	
	\label{tab:Autoencoder}
\end{table*}

\subsubsection{SSIM-Based Loss Function}
The loss function $\mathcal L$ is the effect driver of the neural network's learning, and the loss function in an autoencoder network generally defaults to MSE. However, MSE is prone to lead to a smooth/blur reconstruction which may lose some structural information in the original signal \cite{Zhao2017}. Structural information refers to the knowledge about the structure of objects, e.g. spatially proximate, in the visual scene \cite{Wang2004}. Particularly in this study, structural information mainly refers to the spatial organization of cells in H\&E stained breast cancer {\color{black}histopathology} images. It should be noted that since the multicellular structural information is a key for invasive breast cancer diagnosis \cite{Araújo2017}, it should be learned and maintained in autoencoder's output. To this end, we make use of the SSIM index to compose the loss function for AE's training, i.e.
\begin{eqnarray}
 \mathcal L(\textit x_{i,j},\tilde{\textit x_{i,j}}) = 1-SSIM(\textit x_{i,j},\tilde{\textit x_{i,j}}),
\end{eqnarray}
which facilitate the autoencoder to maintain structural information in image patches. SSIM index is defined as  
\begin{eqnarray}
 SSIM(\textit x,\tilde{\textit x}) = [l(\textit x,\tilde{\textit x})]^{\alpha}\times [c(\textit x,\tilde{\textit x})]^{\beta}\times [s(\textit x,\tilde{\textit x})]^{\gamma},
\end{eqnarray}
where $l(),c(),$ and $s()$ are the luminance comparison function, contrast comparison function, and structural comparison functions, respectively. $\alpha,\beta,$ and $\gamma$ are used to adjust the relative importance of the three components. %, and we follow the setting in the original study and set $\alpha=\beta=\gamma=1$.

\subsubsection{Pattern Learning With True-Normal Patches}

It is noteworthy that in this study, instead of training the network with all training patch $\{\textit x_{11},...,\textit x_{K,l_K}\}$, the autoencoder is trained with only true-normal patches $\{\textit x_{11},...,\textit x_{N,l_{N}}\}$. The motivation behind this innovation is the data-specific property of autoencoder. That is, an autoencoder has low reconstruction errors for samples following training data's generating distribution, while having a large reconstruction error otherwise. Specifically, in our study, the autoencoder learns the common properties and dominant patterns among true-normal patches. Thus, the trained autoencoder is capable of recovering the common content shared by query and true-normal patches. The smaller the residue is, the similar the query patch is to true-normal patches. However, since patches containing invasive tumor have some distinct patterns so that the autoencoder cannot represent well, large construction residue is generated. In other words, the discriminative and contrast patterns in this problem are embedded in autoencoder's residues $\Delta \textit x$, which is quantified by the absolute value of the difference between AE's input and output.
\begin{eqnarray}
 \Delta \textit x = \left|\textit x - \tilde{\textit x} \right| = \left|\textit x -  \mathcal D(\mathcal E(\textit x))\right|.
\end{eqnarray}

To facilitate downstream patch labeling, discriminative patterns embedded in $\Delta \textit x$ are summarized by several compact numerical descriptors. Motivated by the radiomics analysis \cite{Griethuysen2017}, we compute 16 patch-wise first-order statistics to describe the distribution of intensities within AE's residue $\Delta \textit x$, which are energy, minimum, maximum, $10^{th}$ percentile, $90^{th}$ percentile, mean, median, interquartile range, full range, mean absolute deviation, robust mean absolute deviation, variation, skewness, kurtosis, entropy, and histogram uniformity. We denote the obtained numerical feature set by $\{\textit z_{11},...,\textit z_{K,L_{K}}\}$, where $\textit z_{i,j}$ is the description vector of the residue corresponding to patch $\textit x_{i,j}$ and the first $T_N$ elements in the feature set come from the true-normal patches. 

\subsection{Patch Labeling by One-Class SVM}

With the discriminative representation $\{\textit z_{i,j}\}$ generated by a deep convolutional autoencoder, we precede to investigate the mapping function $\mathcal F: \textit z_{i,j} \rightarrow \{-1,+1\}$ from numerical features of true-normal patches. Note that for malignant images with $\mathcal y_{i}=1$, patch labels are not necessarily consistent with image labels; in addition, the number of patches having malignant tumor may be much smaller compared to the quantity of true-normal patches in the training set. As only true-normal patches with their labels are reliable, we cast the problem into the problem of semi-supervised anomaly detection. Intuitively, if one can find regions in the feature space where true-normal patches cluster, patches falling out of the "normal" regions are highly likely to be abnormal.   

Due to the good performance of one-class SVM in medical anomaly detection \cite{Dreiseitl2010,Cichosz2016}, we select one-class SVM to approximate the distribution of true-normal patches for anomaly patch detection. It should be noted that in one-class SVM, normal patches are labeled as +1, which is opposite to the patch labels defined in this study. Hence, based on the features of true-normal patches $\{\textit z_{i,j}:0<i<T_n\}$, a patch label can be obtained using a mapping function 
\begin{eqnarray}
	\mathcal F(\textit z) = -\text{sgn}(\mathcal G(\textit z)=-\text{sgn}\large(\sum_{i=0}^{T_n} \lambda_i \textit k(\textit z_i,\textit z)-\rho \large),	
\end{eqnarray}
where $\rho$ is defined in (\ref{eqn:rho}).

%In this testing phase, with the trained SVM, one image is labels as malignant when at least one of its patches is classified to having cancerous cell clusters. 

\subsection{Malignant Tumor Probability Map Generation}
Now we obtain a labeled training set $\{(z_{i,j},y_{i,j}):0<j\leq l_i,0<i\leq T\}$, where the last $T_M$ samples have labels generated by the one-class SVM. {\color{black} It should be noted that in precision medicine, patches having cancerous tissues definitely (locating far from SVM's hyperplane) and those suspected containing abnormality (residing near SVM's hyperplane) should be distinguished. However, SVM does not provide a posterior probability $p(\textit y_{i,j}|\textit z_{i,j})$ and the resulting labels cannot differentiate these cases. Hence,} for any data sample $\textit z_{i,j}$, we make use of SVM's decision function $\mathcal G(\textit z_{i,j})$ to compute Platt's score for posterior probability approximation \cite{Platt1999}. Platt's score is defined as a sigmoid function on SVM's decision function. That is,
\begin{eqnarray}
	p(\textit y_{i,j}|\textit z_{i,j}) \approx p(\textit y_{i,j}|\mathcal G(\textit z_{i,j})) = \frac{1}{1+e^{(A\mathcal G(\textit z_{i,j})+B)}},
	\label{eqn:platt}
\end{eqnarray}
where $A$ and $B$ are parameters trained using sample labels. 

Examining the Platt's score in (\ref{eqn:platt}), {\color{black}we notice that Platt's score can be implemented by a 1-layer neural network with a sigmoid activation function for two reasons. First, from a theoretical point of view, sigmoid function is a good candidate to generate a probability from a real value (i.e. SVM's decision function $\mathcal G(\textit z_{i,j})$ in this study) because its output can be interpreted as the posterior probability for the most general categorical distribution: Bernoulli distribution. Second, from the application point of view, though there is a "-" sign difference between Platt's score in (\ref{eqn:platt}) and the standard sigmoid function in machine learning, the sign difference can be easily compensated by the trainable parameters in deep learning. Consequently, the recursive optimization of platt's score is realized by training a 1-layer NN with sigmoid activation function.} For the 1-layer NN, input is one-class SVM's decision function. Parameters of the network, a $1\times 1$ transformation matrix $A$ and a bias $B$, can be optimized using training set $\{(\mathcal G(\textit z_i),y_i):0<i\leq T\}$.

%To obtain a probability value indicating how likely a patch contains malignant cell clusters, a shallow NN is applied to the training set $\{(z_i,y_i):0<i\leq T\}$. Table \ref{tab:shallowNN} provides the detailed information about the architecture of the shallow NN.

%\begin{table}	
%\caption{Architecture of the shallow neural network } 
%	\centering
%	\begin{tabular}{ |c | c | c | c | c | }
%	\hline 
%	Layer & Layer type & Filter Size & Activation & Dimension \\ \hline
%	0 & input & &  & $16 \times 1$     \\ \hline
%	1 & fully-connected & $16\times 32 $ & Relu & $32 \times 1$ \\ \hline
%	2 & fully-connected & $32\times 16 $ & Relu & $16 \times 1$ \\ \hline
%	3 & fully-connected & $16\times 1 $ & Sigmoid & $1 \times 1$ \\ \hline
%	\end{tabular}	
%	\label{tab:shallowNN}
%\end{table} 

To infer an image label from obtained patch labels, patch majority voting, where the image label is selected as the most common patch label, is the most common method in literature of breast cancer {\color{black}histopathology} image diagnosis \cite{Spanhol2016b,Araújo2017}. However, in clinical practice, any abnormalities, suspect lesions in particular, should trigger an alarm. Based on this belief, instead of using majority voting, we propose a much stricter rule to combine patch diagnosis results, that is, an image is labeled as benign when all patches are classified as normal.%has the Platt's score larger than 0.5. %its label is predicted using one of the following two criterion:
%\begin{itemize}
%	\item patch majority voting, where the image label is selected as the most common patch label \cite{Spanhol2016b,Araújo2017}. The majority voting is the most common operation to obtain image-level labels from patch-level labels in literature.
%	\item malignant patch detection, where the image is labeled as malignant when one patch has a probability value larger than 50\%. The criterion is more strict and suitable in a medical alert system where any unusual case and suspects should trigger an alarm notification. 
%\end{itemize}
Finally, we proceed to generate a probability map of abnormality for an image. With the obtained probability  $p(\textit y_i|\textit z_i)$, if an image pixel is only contained in one patch, the probability is assigned to the pixel directly. Otherwise (a pixel is in multiple overlapping patches), the probability at the pixel is obtained by averaging probability values of the overlapping patches.

\section{System Implementation and Training Details} \label{sec:implementation}

\subsection{System Implementation and Hyper-parameter Setting}
The proposed method is implemented using python 3.6.6. Each {\color{black}histopathology} image is normalized using the illuminant normalization method \cite{XLi2015}. Then the normalized image is converted to the grayscale version and rescaled to $[0,1]$. The two-step pre-processing mitigates the effect of color variations usually observed in histopathology images on downstream discriminative pattern mining. The autoencoder and the computation of Platt's score are realized using the Keras library which uses tensorFlow as its backend. The One-class SVM is called from the scikit-learn library. 

To compute the SSIM index when training autoencoder, we use the Gaussian filter with size $11\times 11$ to smooth image patches and fuse the luminance comparison function, contrast comparison function, and structural comparison functions with $\alpha=\beta=\gamma=1$ following SSIM's original paper \cite{Wang2004}. %The optimal hyper-parameters of one-class SVM are $\nu = 0.10311$ and $\textit c = 0.00045$ in this study. 

{\color{black}One-class SVM has application-dependent hyper-parameters, $\nu$ and $\textit c$. In this study, we propose the use of the whole training set to select the optimal hyper-parameters.  %Specifically,the mapping function $\mathcal G()$ returned by one-class SVM with a hyper-parameter set of $\{\nu,\textit c\}$ is applied to the whole feature set $\{\textit z_{i,j}: 0<j<l_i,0<i<T\}$ and the resulting labels are $\{\tilde{\textit y}_{i,j}=\mathcal F(\textit z_{i,j})\}$. 
Given a pair of $\nu$ and $\textit c$, after training over true-normal patches, one-class SVM generates a label for each training patch. Though we cannot directly assess patch classification accuracy due to the lack of patch annotation, we can correspond image labels to evaluate the SVM model indirectly. %It is noteworthy that 
Specifically, all patches from normal images are normal and an image is labeled as malignant when at least one of its patches contains cell's malignant growth pattern, i.e. $\max \textit y_{i,j} = \mathcal y_{i},\forall j\in [1,l_i]$. Hence, the one-class SVM's classification accuracy in image level, denoted by $ACC_{img}$, can be measured by 
\begin{eqnarray}
	ACC_{img} = \frac{1}{T}\sum_{i=1}^{T}\delta(\mathcal y_{i}-\max_j \textit y_{i,j}), 
\end{eqnarray}
where $\delta(\dot)$ is the Dirac delta function, i.e. $\delta(x) = 1$ for $x=0$ and $\delta(x) = 0$ otherwise. By comparing all obtained $ACC_{img}$, the one-class SVM with highest image classification accuracy is selected, i.e.
\begin{eqnarray}
	\nu_{opt},\textit c_{opt} = \arg\max_{\nu,\textit c} ACC_{img}. 
\end{eqnarray}}

\subsection{Training Data Augmentation}
For {\color{black}histopathology} images in the training set, each pre-processed image is divided into $35$ patches, each having $256\times 256$ pixels with 30\% overlap at most. Different from conventional data augmentation methods that generates a fixed augmented training set, data augmentation in this study is performed in an online manner with the support of Keras. Specifically, to learn a rotation-invariant AE network, data augmentation operations in this study include patch rotation with an angle randomly drawn from $[0,180)$ degrees, vertical reflection, and horizontal flip. At each learning epoch, transformations with randomly selected parameters among the augmentation operations are generated and applied to original training patches. Then the augmented patches are feed to the network. When the next learning epoch is started, the original training data is once again augmented by applying specified transformations. That is, the number of times each training data is augmented is equal to the number of learning epochs. In this way, the AE network almost never sees two exactly identical training patches, because at each epoch training patches are randomly transformed. For example, with the breast cancer image data set used in this study, we got a basic set of 3750 true-normal patches for AE's training. After 100 epoch learning, the network had seen 375,000 augmented patches in total. We believe the online augmented method helps to fight against network's over-fitting in this study.

\subsection{Network Initialization and Training}
For both the deep convolutional autoencoder and 1-layer neural network for Platt's score, we initialized all weights with zero-mean Xavier uniform random numbers \cite{Glorot2010}. All biases were set to zero. The networks were trained using Adam stochastic optimization with learning rate 0.001, and the exponential decay rates for the first and second moment estimations are set to 0.9 and 0.999. To enforce the autoencoder to learn the {\color{black}dominant} patterns in true-normal patches, the training ran 100 epochs. The 1-layer neural network for Platt's score was trained with 25 epochs. We used 10\% of the training data for validation. The optimal networks for autoencoder and Platt's score were selected based on the proposed SSIM-based loss function and the binary-classification cross-entropy on the validation sets, respectively.

\section{Experimentation} \label{sec:simulation}
In this study, {\color{black}the proposed method is evaluated using the 119 images of the morphologically normal breast tissue and the 140 images of invasive breast cancer in the breast cancer benchmark data set published by the Israel Institute of Technology}\footnote{Please refer to Section II for detailed information about the image set.}. We will first compare image patches reconstructed by autoencoder with loss functions of SSIM and MSE. Then we qualitatively assess the effectiveness of contrast pattern mining by visualizing obtained features and their distribution in a manifold space. Finally image classification and the obtained abnormality maps are examined and compared to prior arts. 

\subsection{Patch Reconstruction Using SSIM and MSE}
{\color{black}The proposed method exploits an autoencoder as a generator of normal patches. A loss function should be selected such that the autoencoder can reconstruct a normal patch as much as possible. In this experiment, we compare reconstructed patches generated by different loss functions, SSIM and MSE, and quantify their effects on normal patch generation. After 100-epoch training over 4165 patches generated from the 119 normal breast tissue images, energies of reconstruction residues over all true-normal patches are calculated and averaged. Specifically for SSIM and MSE based loss functions, the average energies per patch are 194.756 and 219.785, respectively. That is, the SSIM-based function drives the autoencoder to learn more from its inputs. Fig. 3 presents examples of patch reconstructions associated with loss functions of SSIM and MSE, where reconstructed patches associated with MSE is more blurred. }

%\Figure[t!](topskip=0pt, botskip=0pt, midskip=0pt)[width=79mm]{image/MSE-SSIM}
%{{\color{black}Comparison of patch reconstruction using different loss functions, SSIM and MSE. The SSIM-driven reconstructions are sharper than the MSE-driven images. \label{fig:lossFunction}}}

\begin{figure}
	\centering
		\includegraphics[width=3.4in]{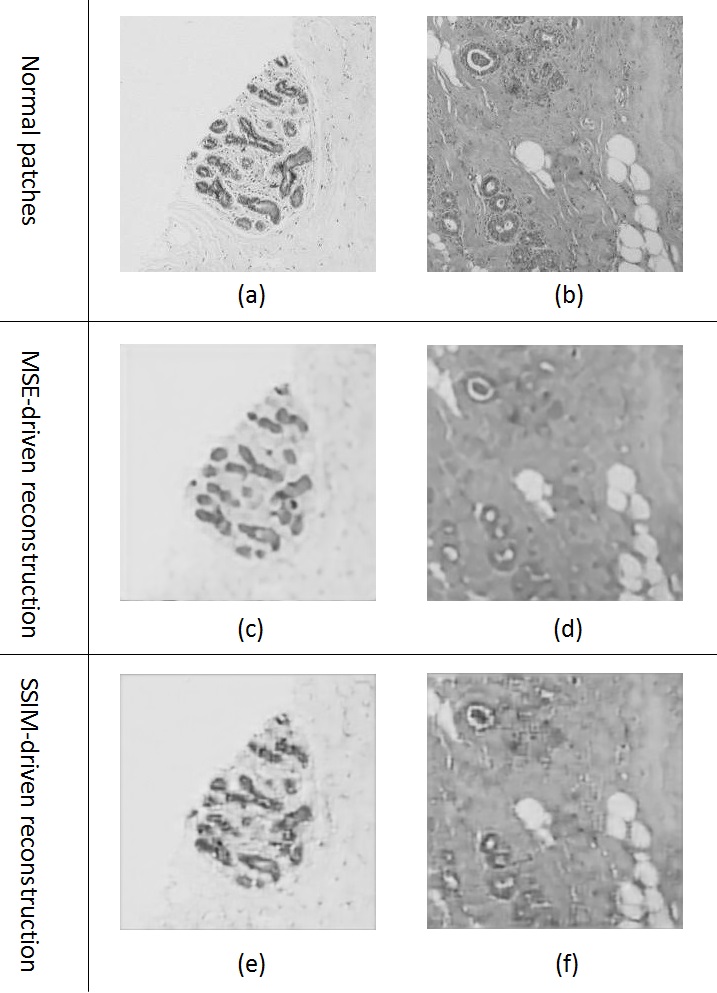}
		\centering	\caption{Comparison of patch reconstruction using different loss functions, SSIM and MSE. The SSIM-driven reconstructions are sharper than the MSE-driven images.}
	\label{fig:lossFunction}
\end{figure}

\subsection{Visualization of Contrast Pattern Mining}

\begin{figure}
	\centering
		\includegraphics[width=3.4in]{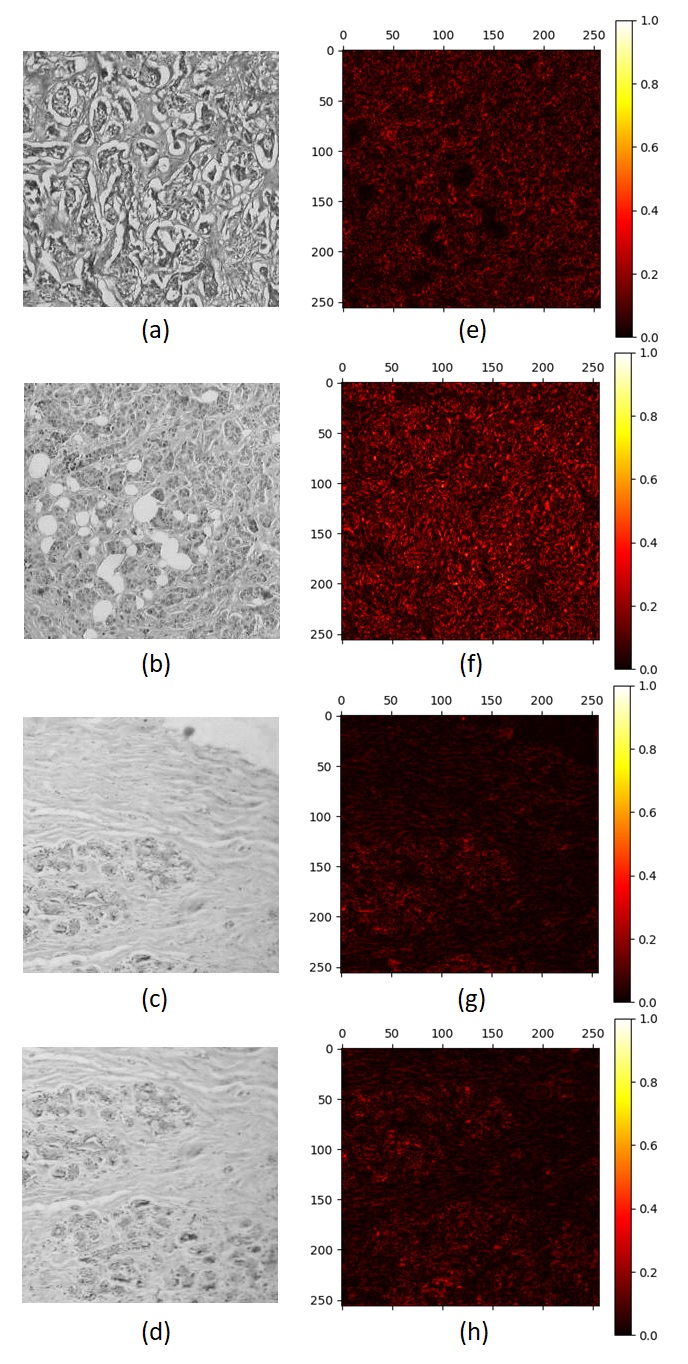}
		\centering	\caption{Examples of the deep autoencoder's inputs ($256\times 256$ grayscale patches from images with a magnification of 40x) (a)-(d) and their reconstruction residues (e)-(h), where brighter values in the residue heatmaps represent large reconstruction errors. The upper two patches from malignant images (a)-(b) contain abnormal cell growth patterns, and the lower two (c)-(d) are extracted from normal breast tissue {\color{black}histopathology} images.}
	\label{fig:Residue}
\end{figure}

A fully convolutional autoencoder net is used to mine the common patterns in normal histopathology image patches. Fig. \ref{fig:Residue} presents several examples of autoencoder's input (in the left column) and their corresponding reconstruction residues (in the right column), where AE's residues are represented as heatmaps for visualization. As shown in the figure, residue images (e)-(f) that correspond to the malignant patches (a)-(b) have brighter values; on contrast, the true-normal patches (c)-(d) have relatively small reconstruction errors (g)-(h). Thanks to the data-specific property of the autoencoder, the {\color{black}dominant} patterns in normal image patches are well summarized, while the abnormal patterns among malignant breast cancer images are maintained in AE's residues.  

%\Figure[t!](topskip=0pt, botskip=0pt, midskip=0pt)[width=80mm]{image/AE_residue}
%{Examples of the deep autoencoder's inputs ($256\times 256$ grayscale patches from images with a magnification of 40x) (a)-(d) and their reconstruction residues (e)-(h), where brighter values in the residue heatmaps represent large reconstruction errors. The upper two patches from malignant images (a)-(b) contain abnormal cell growth patterns, and the lower two (c)-(d) are extracted from normal breast tissue {\color{black}histopathology} images. \label{fig:Residue}}

\begin{figure}
	\centering
		\includegraphics[width=3.0in]{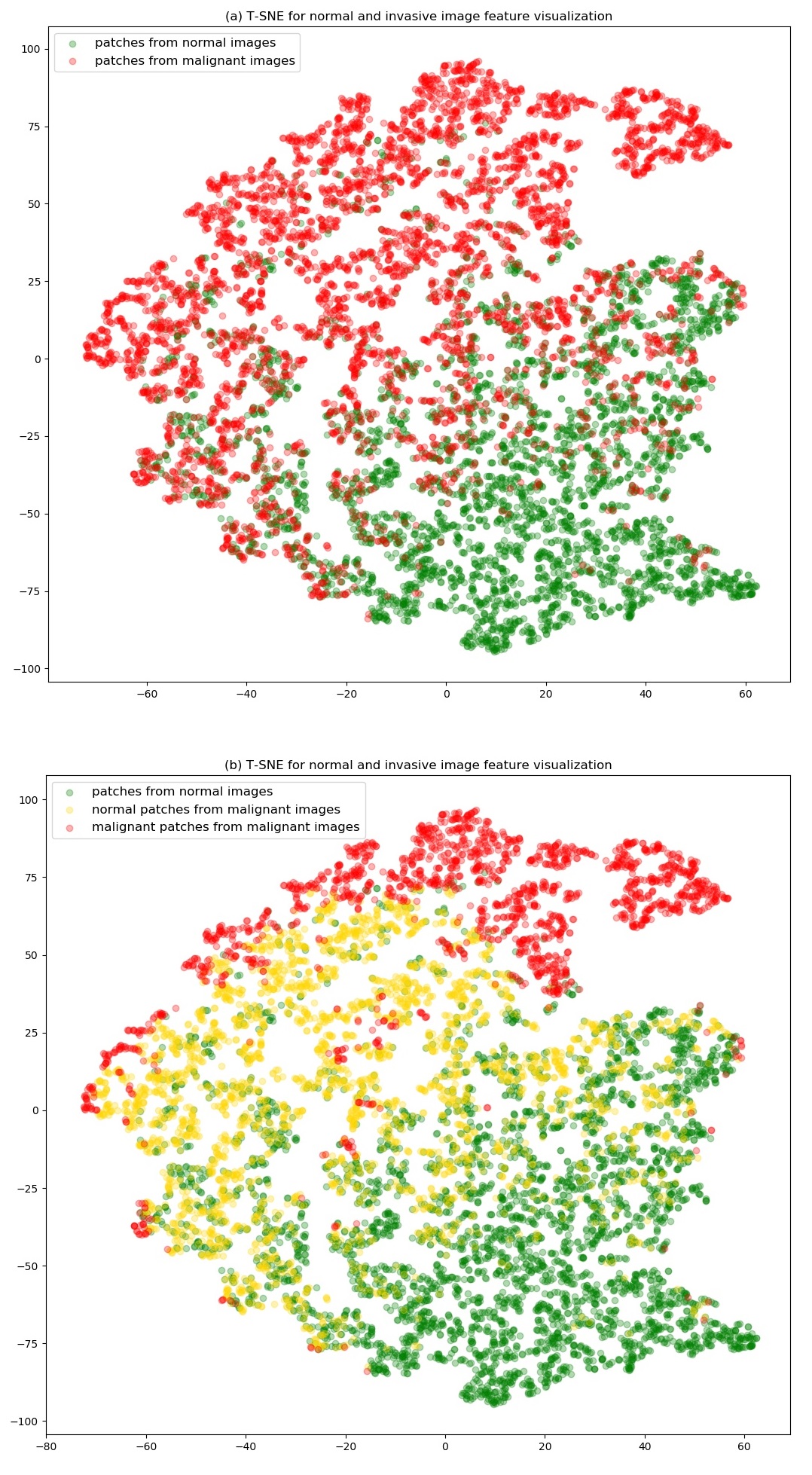}
		\centering	\caption{Feature set visualization via T-SNE. (a) Some patches from malignant images overlap with true-normal patches in the low-dimensional T-SNE domain. (b) The yellow samples represent patches that are extracted from malignant images and classified as normal by the one-class SVM.}
	\label{fig:TSNE}
\end{figure}

To visualize the distribution of learnt contrast patterns, we project the obtained high dimensional feature set into 2-D domain via T-SNE \cite{Maaten2008} and illustrated in Fig. \ref{fig:TSNE}(a). A green sample is associated with a true-normal patch and a red sample represents a patch from a malignant image. From the figure, more than half red samples share different characteristics from green samples. Fig \ref{fig:TSNE}(b) visualizes the performance of one-class SVM, where green sample corresponds to a true-normal patch, while yellow and red samples are associated with patches from malignant images in the 2-D T-SNE domain. The difference is that yellow samples are classified as normal by the one-class SVM, while red data represent patches that are labeled as containing malignant cell clusters. 

%\Figure[t!](topskip=0pt, botskip=0pt, midskip=0pt)[width=70mm]{image/TSNE}
%{Feature set visualization via T-SNE. (a) Some patches from malignant images overlap with true-normal patches in the low-dimensional T-SNE domain. (b) The yellow samples represent patches that are extracted from malignant images and classified as normal by the one-class SVM.\label{fig:TSNE}}

%\subsection{Sensitivity of Parameter Setting}
\subsection{Breast Cancer Image Classification}

\subsubsection{Evaluation Protocol}
{\color{black} To evaluate the proposed method, stratified 10-fold cross-validation is performed. Specifically, the image set is randomly partitioned into 10 equal-size folds, where each fold contains roughly the same proportions of normal and malignant labels. The cross-validation is repeated 10 times where each fold is used as the test set once and images in remaining 9 folds are used as training data. Then the obtained 10 diagnosis results are averaged to estimate classification performance. In each round of cross-validation,} images in the training set are processed as described in the section of data augmentation. In the testing phase, image patches are extracted every 16 pixels, i.e. the centers of two patches may be only 16 pixels apart in an image. The distance of 16 pixels is a trade-off between generating a fine probability map and maintaining computation efficiency. 

In this experiment, we first perform a quantitative evaluation on image classification. Particularly, to measure image classification performance, we use the most common medical diagnosis assessments, which include classification accuracy $ACC\in [0,1]$, F-measure score $F_1\in [0,1]$, positive/negative likelihood ratios $LR+\in [1,\infty)$ and $LR-\in[0,1]$, and diagnostic odds ratio $DOR\in[1,\infty)$. $ACC$ is one of the most common classification performance measurements. It represents the proportion of accurate diagnoses, but it is impacted by disease prevalence. $F_1$ is the harmonic average of the precision and sensitivity and $F_1 = 1$ corresponds to a perfect binary classification. Likelihood ratios use the sensitivity and specificity of the test to determine its diagnostic performance. They are believed as good institutions of AUC when ROC analysis is infeasible. $DOR$ combines sensitivity and specificity and equals to the ratio of positive and negative likelihood ratio. Among the five measurements, $F_1$ score, likelihood ratio, and $DOR$ are independent of test prevalence, with higher values indicating a better discriminative performance.

Since the breast cancer image set does not delineate the specific locations of malignant cell clusters, we perform a qualitative assessment of the obtained probability maps by comparing it to abnormality regions derived by malignant cell growth. 
 %Experimental evaluation is repeated 10 times to get a reliable assessment. 

\subsubsection{Other Approaches}
{\color{black} To the best of our understanding, the proposed method constitutes the first attempt in literature of breast cancer diagnosis to infer locations of abnormalities from image labels. Since there is no such breast cancer diagnosis study in literature, we compare the performance of the proposed method to the latest patch-based deep-learning breast cancer histopathology image classification methods proposed by Spanhol et. al \cite{Spanhol2016b} and Araujo et.al \cite{Araújo2017}. Spanhol's method divides an image into $64\times 64$ image patches and uses an 8-layer convolutional neural network to classify image patches. Then three fusion rules, majority voting, malignant patch detection (i.e. maximum probability), and sum of probability, are used to obtain the final image classification.} Araujo's method divides images into $512\times 512$ patches and enforces training-patch labels consistent with image labels. Based on training patches and their newly-assigned labels, a 13-layer convolutional neural net is trained and used to classify unseen image patches. The final image classification is also achieved by combining all inferred patch labels using one of the three rules used in Spanhol's method. {\color{black}Noted that in both studies, malignant patch detection is reported to achieve worse performance than the other two fusing rules for image diagnosis. However, based on the belief that in a medical alert system, any suspected alterations should trigger an alarm, we select the fusing rule of malignant patch detection for both image classification methods in this comparison experiment.}

\subsubsection{Results and Discussions}
%\subsubsection{Image Classification}

%\Figure[t!](topskip=0pt, botskip=0pt, midskip=0pt)[width=160mm]{image/P_map}{Examples of breast tissue images and their corresponding abnormality probability maps, where probability value greater than 0.5 indicates abnormalities in this study. In images in the second row, abnormality regions derived by malignant cell growth were delineated red.\label{fig:P_map}}

\begin{table}	
{\color{black}\caption{Image-level classification evaluation. The sign $^{\star}$ indicates that the performance difference between the prior method and the proposed method is of statistical significance at the 5\% significance level. } }
	\centering
	{\color{black}\begin{tabular}{ |c | c | c | c | }
	\hline 
	 & Spanhol's \cite{Spanhol2016b}  & Araujo's \cite{Araújo2017} & proposed \\ \hline	
	$ACC$ & 0.700$^{\star}$ & 0.710 & 0.760 \\ \hline
	$F_1$ & 0.766$^{\star}$ & 0.763$^{\star}$ & 0.777 \\ \hline
	$LR+$ & 1.540$^{\star}$ &  1.554$^{\star}$ & 2.645 \\ \hline
	$LR-$ & 0.229$^{\star}$ & 0.281$^{\star}$ & 0.304 \\ \hline
	$DOR$ & 8.563 & 8.132 & 12.876 \\ \hline
	\end{tabular}	}
	\label{tab:classification}
\end{table}

\begin{figure*}
	\centering
		\includegraphics[width=6.8in]{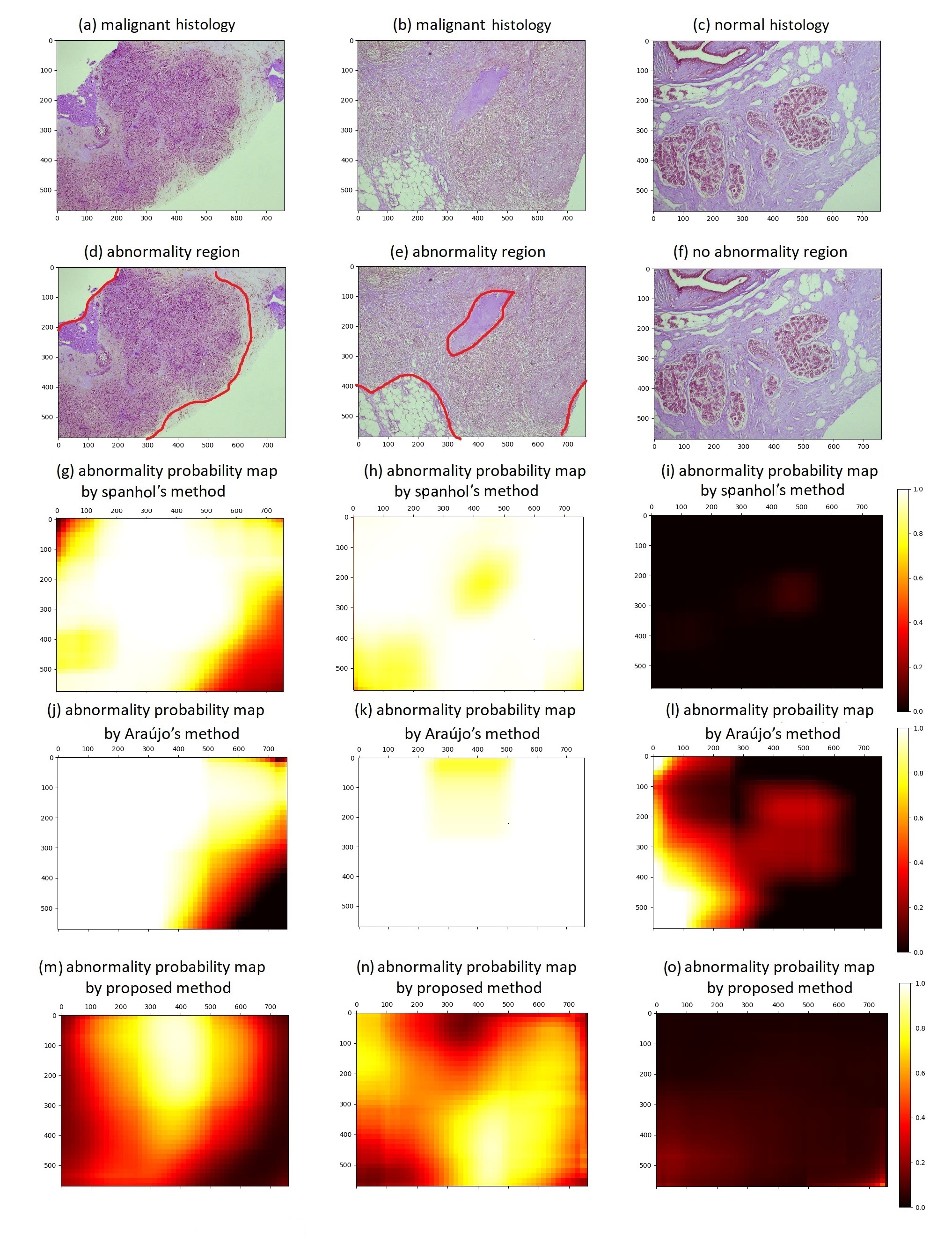}
		\centering	\caption{Examples of breast tissue images and their corresponding abnormality probability maps, where probability value greater than 0.5 indicates abnormalities in this study. In images in the second row, abnormality regions derived by malignant cell growth were delineated red.}
	\label{fig:P_map}
\end{figure*}

\begin{table*}	
{\color{black}\caption{Comparison summary of examined methods } }
	\centering
	{\color{black}\begin{tabular}{ |c | c | c | c | }
	\hline 
	 & Spanhol's method \cite{Spanhol2016b}  & Araujo's method \cite{Araújo2017} & proposed method \\ \hline	
	Training procedure & 1 end-to-end training & 1 end-to-end training & 3 training phases \\ \hline
	Practicality & no & no & yes \\ \hline
	Generalizability & less generalizable & less generalizable & generalizable \\ \hline
	classification accuracy & fair &  fair & better \\ \hline
	self-interpretability & no & no & yes\\ \hline
	\end{tabular}}	
	\label{tab:methodComparison}
\end{table*}

Table \ref{tab:classification} lists the image classification performance. The sign $^{\star}$ indicates that the performance difference between the prior method and the proposed method is of statistical significance at the 5\% significance level. %Two observations are obtained. First, the proposed method outperforms the state-of-the-art Araujo's method except the negative likelihood ratio. This is because the proposed method tries to discover the contrast patterns between normal breast tissue and abnormalities, which leads to a discriminative diagnosis. On contrast, Araujo's method enforces training patch labels consistent with image labels, which are not always hold in practice. Second, the performance of Araujo's method with malignant patch detection is slightly worse than the classification results of the original method with patch majority voting. Though patch majority voting is the most common operation to obtain image labels from patch labels in literature \cite{Spanhol2016b,Araújo2017}, in a medical alert system, we believe that any suspected alterations of normal tissue histomorphology should trigger an alarm. With this strict patch-classification fusion method, any normal images containing one miss-classification patch goes to the malignant category.
%	\item malignant patch detection, where the image is labeled as malignant when one patch has a probability value larger than 50\%. The criterion is more strict and suitable in a medical alert system where any unusual case and suspects should trigger an alarm notification.
{\color{black}The better performance of the proposed method is mainly contributed by its practicality and generalizability in contrast pattern mining. First, the minimum prior knowledge to train the proposed method is WSI label which can be easily acquired in clinic practice. But this information is not enough to train Spanhol's method \cite{Spanhol2016b} and Araujo's method \cite{Araújo2017}; instead, patch-label or even pixel-wise annotation is required. It should be noted that these prior deep models have more than ten thousands of trainable parameters. Acquisition of sufficient amount of well-labeled data for these models is fairly prohibitive, if not impossible, in practice due to the expensive and time-consuming properties of pathological annotations. To address the shortage of well-labeled training patches for supervised learning, Araujo's method makes an assumption that patch-labels are identical to their image labels \cite{Araújo2017}. However, this assumption hardly holds in practice because a tumor usually takes 0.01\%-70\% (median 2\%) areas of a WSI image \cite{Google2017}. The less practical requirement/assumption on training data in prior arts limits their diagnosis performance in practice. Second, in pathology, normal breast tissues usually share common histological patterns, whereas structures and patterns of malignant cell clusters are heterogeneous. Consequently, quantification of the normal patterns is relatively feasible, but representation learning among histological irregularities is more challenging. In prior studies, cell's abnormal patterns are usually learnt directly from training samples. However, due to the limited amount of training data, variations in histological abnormalities may not be fully represented. As a result, the generalizability of these deep diagnosis models to unseen malignant cancer images is still in question. To overcome the challenge of abnormality representation in digital pathology, the proposed method learns the common patterns in normal breast images first and diagnoses malignant cells by similarity of these cells and their growth patterns to normal ones. In this way, detection of histological abnormalities is simplified to identification of common patterns in normal breast images. Consequently, the proposed method is less dependent on specific malignant image samples and can generalize well.}
Examples of the obtained probability maps with their corresponding H\&E images are demonstrated in Fig. \ref{fig:P_map}. Abnormality regions derived by malignant cell growth in the query images were delineated by our senior pathologist and highlighted in images at the second row. A probability map is presented in the form of a heat-map where bright pixels represent high probabilities of abnormalities. It provides an insight and verification of the image diagnosis result by inferring locations of abnormalities in an image. In this sense, it even conveys more information compared to the classification result itself. Since Spanhol's's approach and Araujo's method are also based on patch processing, as a comparison, the obtained patch-level probabilities are used to form the corresponding probability maps following the method proposed in this study. {\color{black} As shown in the figure, the two prior methods are prone to yield large probabilities in background areas of invasive breast cancer histopathology images.} %Again, the superiority of our method is contributed by the effectiveness of the proposed discriminative pattern learning method. 

In summary, {\color{black}the advantages of the proposed method are contributed by its practicality, generalizability, and self-interpretability. First, the minimal prior knowledge on whose-slide-imaging (WSI) labels for system training is easily acquired in clinic practice. Second, the proposed method detects discriminative patterns in images in unsupervised manner. Because the method is less dependent on specific malignant image samples, it generalizes well on unseen images. Third, the obtained probability map infers locations of abnormalities in an image. The insightful information explains the final diagnosis result and helps pathologists to verify the diagnosis reasoning. Table 4 summarizes a comprehensive comparison of examined methods.} The major limitation of this study is the size of the experimental image set and the absence of the external validation group. However, the carefully designed experimentation and the involvement of pathologist's expertise in this study support the reliability of the obtained results. In addition, the public-accessibility of the experimental image set facilitates other scholars to reproduce our study.

\section{Conclusion}\label{sec:conclusion}
In this study, we presented a discriminative pattern mining approach for invasive carcinoma diagnosis in routine H\&E stained breast tissue {\color{black}histopathology} images. By learning contrast patterns between normal and malignant breast cancer images, the proposed method was capable to identify suspected regions of malignant cell clusters in an image. The evaluation was conducted on a public {\color{black}histopathology} image set and experimentation demonstrated that the proposed method outperformed prior arts. Particularly, the superiority of the proposed method was its practicality, generalizability, and self-interpretability. The obtained probability map would facilitate a better understanding of the proposed pattern mining and diagnosis solution.

{\color{black}In this study, heterogeneity of histological abnormalities posed a big challenge in pattern mining. we noted that there was still a large room to improve the diagnosis performance by investigating more efficient pattern mining methods. On the other hand, though we tried to tackle the problem of self-interpretability in machine learning and proposed a diagnosis system which was capable to generate visualized information to support and verify its decision, the black-box property of deep learning in terms of data representation was still less-touched. Following the work of CAM \cite{CAM2016} and Grad-CAM \cite{Grad-cam2017}, we would investigate how to interpret the internal reasoning of a deep diagnosis system in future.}

%\begin{thebibliography}{00}
%\end{thebibliography}

\ifCLASSOPTIONcaptionsoff
  \newpage
\fi

\bibliographystyle{IEEEbib}
%\bibliography{strings,refs}
%\onecolumn
%\bibliography{refs}


%\end{IEEEbiography}


%\newpage



%\vfill


\end{document}